\documentclass[letterpaper,twocolumn,10pt]{article}
\usepackage{usenix2019_v3}

\usepackage{graphicx}
\usepackage{booktabs} 
\usepackage{xspace}
\usepackage{hyperref}
\usepackage{url}
\usepackage{etoolbox} 

\usepackage{color}
\usepackage{graphicx}
\usepackage{alltt}
\usepackage{lineno}
\usepackage{proof}
\usepackage{verbatim}
\usepackage{xcolor}
\usepackage{tikz}
\usepackage{xspace}
\usepackage{fancyvrb}
\usepackage{xfp} 
\usepackage{subcaption}
\usepackage{xspace}
\usepackage{multirow}

\usepackage{caption}
\newcommand{\rom}[1]{\uppercase\expandafter{\romannumeral #1\relax}}

\newcommand{\system}[0]{\emph{Genesis}\xspace}
\definecolor{codegreen}{rgb}{0,0.6,0}
\definecolor{codegray}{rgb}{0.5,0.5,0.5}
\definecolor{codepurple}{rgb}{0.58,0,0.82}
\definecolor{backcolour}{rgb}{0.95,0.95,0.92}
\definecolor{amber}{rgb}{1.0, 0.49, 0.0}
\usepackage{listings}

\definecolor{sampldark}{RGB}{72, 71, 135}
\definecolor{uwpurp}{RGB}{51, 0, 111}
\definecolor{uwpurplight}{RGB}{94, 0, 204}

\usepackage{graphics}

\usepackage{todonotes}

\newcommand{\rnum}[1]{\uppercase\expandafter{\romannumeral #1\relax}}

\newcommand*\circled[1]{\tikz[baseline=(char.base)]{
            \node[shape=circle,fill,inner sep=2pt] (char) {\textcolor{white}{#1}};}}

\usepackage{ifthen}
\newcommand{\showcomments}{yes}
\newcommand\zhi[1]{
    \ifthenelse{\equal{\showcomments}{yes}}{{\color{red} [zhi: #1]}}{\ignorespaces}
}
\newcommand\cody[1]{
    \ifthenelse{\equal{\showcomments}{yes}}{{\color{blue} [cody: #1]}}{\ignorespaces}
}
\newcommand\yida[1]{
    \ifthenelse{\equal{\showcomments}{yes}}{{\color{cyan} [Yida: #1]}}{\ignorespaces}
}
\newcommand\sean[1]{
    \ifthenelse{\equal{\showcomments}{yes}}{{\color{purple} [Sean: #1]}}{\ignorespaces}
}

\newcommand\hide[1]{}

\usepackage{stfloats}

\usepackage{lipsum}

\begin{document}

\date{}

\title{Bring Your Own Codegen to Deep Learning Compiler}

\author{
     {\rm Zhi Chen\thanks{Equal contribution}}\\
     chzhi@amazon.com \\
     Amazon Web Services, Inc
 \and
     {\rm Cody Hao Yu\footnotemark[1]}\\
     hyuz@amazon.com \\
     Amazon Web Services, Inc
  \and
     {\rm Trevor Morris}\\
     trevmorr@amazon.com \\
     Amazon Web Services, Inc
  \and
     {\rm Jorn Tuyls}\\
     jornt@xilinx.com \\
     Xilinx
   \and
     {\rm Yi-Hsiang Lai\thanks{This work is done when Yi-Hsiang Lai was an intern at Amazon.}}\\
     yl2666@cornell.edu \\
     Cornell University
  \and
     {\rm Jared Roesch}\\
     jroesch@octoml.ai \\
     OctoML, Inc
   \and
     {\rm Elliott Delaye}\\
     elliott@xilinx.com \\
     Xilinx
  \and
     {\rm Vin Sharma}\\
     vinarm@amazon.com \\
     Amazon Web Services, Inc
  \and
     {\rm Yida Wang}\\
     wangyida@amazon.com \\
     Amazon Web Services, Inc
} 


\maketitle

\begin{abstract}
Deep neural networks (DNNs) have been ubiquitously applied in many applications, and accelerators are emerged as an enabler to support the fast and efficient inference tasks of these applications.
However, to achieve high model coverage with high performance, each accelerator vendor has to develop a full compiler stack to ingest, optimize, and execute the DNNs.
This poses significant challenges in the development and maintenance of the software stack.
In addition, the vendors have to contiguously update their hardware and/or software to cope with the rapid evolution of the DNN model architectures and operators.
To address these issues, this paper proposes an open source framework that enables users to only concentrate on the development of their proprietary code generation tools by reusing as many as possible components in the existing deep learning compilers.
Our framework provides users flexible and easy-to-use interfaces to partition their models into segments that can be executed on ``the best'' processors to take advantage of the powerful computation capability of accelerators.
Our case study shows that our framework has been deployed in multiple commercial vendors' compiler stacks with only a few thousand lines of code.

\end{abstract}

\section{Introduction} \label{sec:intro}
Deep neural networks (DNNs) have recently achieved tremendous breakthroughs in various domains, such as autonomous driving~\cite{cordts2016cityscapes}, speech recognition~\cite{graves2013speech}, and machine translation~\cite{devlin2018bert}, etc. 
The achievements come at the expense of high computation and memory requirements to handle the potentially high dimensional arrays (a.k.a. tensors) that typically require intense computing resources.
However, executing these applications on the cloud using powerful servers poses strict requirements on network latency and data privacy when the data is collected and the results are needed on the user end.
To overcome the issues, edge computing has been emerged as a popular paradigm for processing DNNs, i.e., executing the model directly at the edge, so that the transmission is omitted and the data is preserved locally.
To cope with the intensive computation, one popular solution is to use the special-purpose hardware accelerator which aims to achieve peak performance on DNN model inference tasks with much lower cost and power budget.
For instance, various edge hardware accelerators, such as Nvidia Jetson family \cite{nvidianano, nvidiaxavier}, Xilinx Vitis AI \cite{xilinxvitis}, ARM Ethos-N \cite{armethosn}, and Google’s Edge Tensor Processing Unit (TPU) \cite{edgetpu}, have emerged in the past few years to support AI applications at the edge.

A large body of deep learning models, particularly image classification ones, are designed with a stack of compute-intensive blocks that consist of convolution and matrix multiplication operators.
These operators are essentially the combination of a determined order of multiplication and addition.
The hardware accelerators are customized for these operators, and thus are promising in handling deep learning models efficiently.
In order to support the entire model, however, only handling the compute-intensive operators in high performance is insufficient or even impractical. For example, more advanced deep learning models may contain complex operators (e.g., data processing) and network architectures (e.g., control flow) that are notoriously difficult for deep learning accelerators to process due to the presence of non-linearity.
To make these deep learning models still functional, each vendor has to either develop more sophisticated software approaches or completely redesign the hardware, leading to the prohibitively high engineering cost.

In addition, accelerator vendors currently have to develop a full compiler stack to accept the model representation, perform a series of optimizations, generate the hardware compatible code, and finally execute the model inference.
Some of these tasks can be shared by many vendors' software toolchains, being orthogonal to the vendor-specific compute-intensive kernel optimization.
Unfortunately, due to missing a unified framework, different vendors have to duplicate the same effort from time to time, significantly distracting them from only focusing on their secret sauce of kernel optimization and high performance code generation.
To summarize, existing research and development of the compiler stack for edge accelerators have been hindered for multiple reasons:

\noindent\textbf{Challenge~1: Failing to execute.} The evolution of deep learning models occurs rapidly which may lead to significant changes in the model architecture over time, e.g., AlexNet~\cite{Krizhevsky2012nipsalexnet} to ResNet~\cite{hecvpr2016resnet} (with residual connection) to Inception~\cite{szegedy2015going} (with representational bottleneck) to SSD~\cite{liu2016ssd} and Mask R-CNN~\cite{maskrcnn} (with control flow). It is non-trivial for accelerators to keep the pace of the emergence of new models without dramatic architectural modification. Therefore, existing accelerators may completely fail to support models with complicated structures.

\noindent\textbf{Challenge~2: Inefficient to execute.} It is also non-trivial to achieve high performance for a model even if its architecture is a simple dataflow, because it may contain a portion of operators that are not executed in a determined order but with complex control logic.
For example, non-maximum suppression (NMS) is widely adopted in object detection models to select a bounding box out of the overlapping ones.
These operators are not a good fit to accelerators that are designed for compute-intensive workloads.
Hence, they could be at most insufficiently supported, if not impossible.

\noindent\textbf{Challenge~3: Demanding to execute.} Even if all operators in a model can be well supported, each vendor still has to spend a considerable amount of engineering efforts to develop a full compiler stack to ingest, optimize, and compile the model due to the lack of a unified framework that allows them to reuse the platform-agnostic optimization techniques such as standard compiler optimizations (e.g., dead code elimination, common sub-expression elimination, constant folding, etc) and deep learning model specific optimizations (e.g., layout transformation~\cite{liu2019optimizing}).

    

To address these challenges, in this paper, we present a unified framework to facilitate hardware vendors to only focus on the proprietary codegen while leveraging the existing standard pipeline to handle the DNNs, which substantially shortens the software stack development cycles.
Specifically, our framework decouples a deep learning compiler (DLC) to two parts -- the shared part and the accelerator-exclusive part.
The shared part can be treated as a DLC for general processors such as CPUs and GPUs.
Vendors can contribute the features required to execute new emerged models to the shared part and make them available on general devices to benefit each other, so \textit{Challenge~1} is resolved.

While the shared part of the DLC is capable of executing all new emerged models on general devices, the accelerator-exclusive part can be maintained by each vendor.
Our framework offers an easy-to-use interface for users to annotate and partition the model strategically and offload the supported and efficient segments to the accelerator while leaving the rest of it on the host.
The partitioning framework is a powerful enabler for users to adapt new models and/or new operators without redesigning the hardware.
As a result, \textit{Challenge~2} is overcome.




In addition, our framework provides the accessibility to the widely used DLC optimizations to deal with \textit{Challenge~3}.
The system paves the way for accelerator vendors to only concentrate on the hardware-specific optimizations and code generation.
This allows vendors to fully exploit the available techniques for quick compiler development with state-of-the-art performance optimization while not revealing their intellectual property.
We have integrated the tool chains of several hardware vendors with the framework to enjoy all optimizations provided by the framework using only $\sim$2k LOC on average (see \autoref{sec:study} for details).
This notably reduces the compiler development time.

To the best of our knowledge, this is the first practical solution that provides a unified, customizable compilation framework for users to integrate their own accelerator code generation to a DLC, and it has been used by multiple commercial accelerators.
In summary, this paper makes the following contributions.

\begin{itemize}
    \item We propose a unified framework to allow different hardware accelerator vendors to obtain as many as possible hardware-agnostic optimizations for free by integrating their codegen tools in a plug-and-play manner.
    \item We provide flexible interfaces for developers to 1) annotate and partition a computational graph with various strategies, 2) apply hardware-specific optimizations on the partitioned graphs to further improve the performance.
    \item We conduct a number of case studies using multiple popular edge accelerators to demonstrate the different ways of codegen integration with on average only $\sim$2K LOC, largely saving the engineering efforts and time to the market.
    \item The proposed framework has been adopted by several production edge accelerators' compilation pipelines to alleviate the development efforts of the full software stack.
\end{itemize}


The rest of the paper is organized as follows. \autoref{sec:motivation} details the challenges of deep learning compilers for embracing deep learning accelerators to motivate this work. \autoref{sec:design} presents the design and implementation of our framework. \autoref{sec:study} provides the case studies of the integration of multiple popular accelerators using our framework and demonstrates the performance gain. Related work is discussed in \autoref{sec:relw}. \autoref{sec:concl} concludes the paper.

\section{Motivation}
\label{sec:motivation}

\begin{figure}[t]
	\centering
    \includegraphics[width=0.75\linewidth]{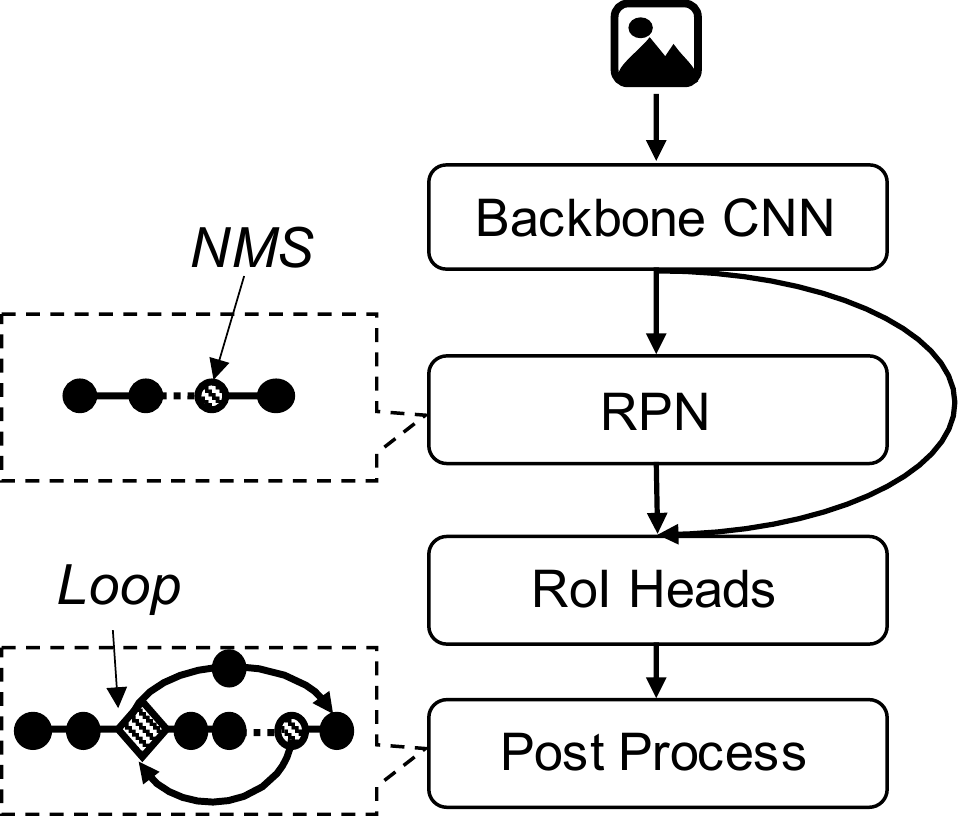}
    \caption{A general R-CNN network architecture. The 
Non-maximum Suppression (NMS) in Region Proposal Network (RPN) and the loop (control flow) in post process are non-trivial for accelerators to support.}
	\label{fig:rcnn}
\end{figure}

In this section, we introduce an example to motivate the system.
As we mentioned in \autoref{sec:intro}, modern deep learning models may include new operators that are non-trivial for accelerators to execute.
One example is R-CNN (region-based convolutional neural network)~\cite{girshick2014rich}, a modern object detection deep learning model, as shown in \autoref{fig:rcnn}.
R-CNN leverages a Region Proposal Network (RPN) to extract a set of proposals (i.e., image regions which likely contain an object) from images based on the selective search~\cite{uijlings2013selective}.
Since classifying a limited number of regions instead of all possible regions in an image is very efficient, the R-CNN family is widely adopted in recent years.

The RPN usually involves a non-maximum suppression (NMS) operation.
This operation filters the proposals (e.g. bounding boxes) that possibly cover the same object to reduce redundancy. It first sorts all proposals based on their scores, then checks each of them against the anchor box, and finally discards the ones that have Intersection over Union (IoU) with the anchor box above a certain threshold.
Sorting may not be supported by most deep learning accelerators that mainly perform tensor computations.
In addition, R-CNN implementation variants (e.g., torchvision~\cite{fb2016torchvision}) normally include loop structures, as shown in the post process module of \autoref{fig:rcnn}.
The model graph with loop structures implies a control flow with \texttt{IF} nodes, which may not be easily supported by existing accelerators leveraging dataflow-based architectures.
Although the unsupported operators and structures only occupy an insignificant portion of the R-CNN, it is cumbersome for an accelerator-specific compiler to partition the model graph, generate code, and deal with runtime graph execution and data transfer.

Existing solutions~\cite{onnxruntime,rotem2018glow} are either not flexible enough for hardware vendors to integrate their compilation toolchain or lack of details. We propose our solution in the next section that provides a comprehensive and flexible solution for hardware vendors.
As we will present in \autoref{sec:study}, our framework has been adopted by a number of hardware vendors in production as a critical component of their compilation toolchain.
\section{Framework Design and Implementation}
\label{sec:design}
\begin{figure*}[tbh]
	\centering
    \includegraphics[width=0.95\linewidth]{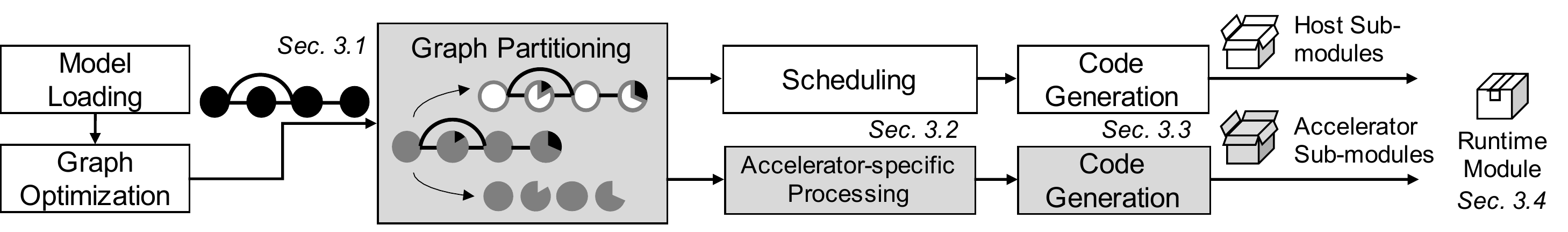}
    \caption{The \system compilation overview. The rectangles in grey are major components in \system. The deep learning model graph represents the general R-CNN. Each dot in the graph denotes a R-CNN module in \autoref{fig:rcnn}.}
	\label{fig:system_overview}
\end{figure*}

This section describes the framework design with the considerations of challenges in \autoref{sec:motivation}.
\autoref{fig:system_overview} presents the compilation flow of our framework.
Note that it is built on top of a deep learning compiler, the system modules in white are existing components in the compiler.
These common deep learning compiler modules can be developed in one place and reused by all accelerators.
On the other hand, the components in grey are the modules related to hardware accelerators, and they can be customized for each accelerator.

We now use the motivational example from the previous section to walk through the process at a high level.
As can be seen from the figure, our framework first loads a deep learning model and constructs the model graph in an intermediate representation (IR).
Then, the graph optimization performs a series of hardware-independent graph-level optimizations, such as operator simplification, constant folding, etc.

Afterwards, the framework partitions the graph to two parts -- ``\textit{host}'' and ``\textit{accelerator}'' based on the given rules(\autoref{sec:design_partition}).
The accelerator part is composed of one or more subgraphs.
Each subgraph will go through the specialized compilation flow, including accelerator-specific optimization (\autoref{sec:design_backend_opt}) and code generation (\autoref{sec:design_codegen}), to generate an executable kernel for runtime offloading.
In this example, four subgraphs (kernels) are sent to the accelerator for efficient processing: the entire backbone CNN; RoI (Region of Interest) head modules; most operators in the RPN module except for the NMS; most operators in the post-processing module except for the loop that contains control flow.

On the other hand, the host part is almost the same as the original graph, with the four accelerator subgraphs replaced by external function calls to invoke the corresponding kernel at runtime.
The remaining operators will go through the normal compilation flow that performs scheduling and code generation for the host, which is usually a general-purpose CPU or GPU.
Finally, both compiled host and accelerator modules are integrated to a single runtime module.
The execution of the runtime module is detailed in \autoref{sec:design_runtime}.

In the rest of this section, we introduce each system component along with the design options and insights in detail.

\subsection{Graph Partitioning}
\label{sec:design_partition}
Deep learning accelerators are usually designed to accelerate the compute-intensive operations, which are composed of massive regular computations, such as multiplication and accumulations (MACs).
While these operators account for the majority portion of a DNN, failing to execute other operators (efficiently) on the accelerator would cause significant performance degradation or even failures in execution.

In order to guarantee the successful and performant execution, we design a partitioning module for users to flexibly cut their model graphs into various regions/subgraphs.
Only the accelerator friendly regions are offloaded, and the rest of the graph is left to the host. 
Since modern deep learning compilers~\cite{tvm, ragan2013halide, baghdadi2019tiramisu, rotem2018glow, tensorcomprehension} generally feature multi-level IRs to better optimize the model with different analysis techniques and information, we need to make a design decision on which IR level we should partition and offload the subgraphs.


In the framework, we choose to partition and offload the graph at the high-level IR that includes the operator information, such as operator name (e.g., \texttt{Conv2D}) and its attributes (e.g., stride, padding, and dilation) due to the following reasons.
First, some hardware vendors handcraft a kernel library with limited inter-operator optimizations.
In this case, the only information they need is the name and attributes to map each operator to the corresponding kernel instead of the hardware information incorporated by the low-level IR, so high-level IR is ideal for them to customize the code generation.
Second, even other hardware vendors prefer to generate the code for accelerator-specific instructions, high-level IR is also more flexible for them to connect the desired low-level IR.
Vendors can choose to use their own IR or the builtin low-level IR in existing deep learning compilers.


In the rest of this subsection, we present the implementation details of graph partitioning using an example in \autoref{fig:partition}.

\subsubsection{Pattern-based Grouping}
\label{sec:design_partition_pattern}
Many deep learning specific hardware accelerators have powerful instructions to execute a sequence of operators with peak performance.
For example, the sequence of \texttt{Conv2D}, \texttt{Add} and \texttt{ReLU} can usually be mapped to a single instruction to minimize overheads in dealing with intermediate results.
As a result, it is a common requirement from many hardware vendors to employ a pattern matching algorithm to match a sequence of IR nodes and replace them with the composite instruction.
This, however, leads to tedious and redundant work for each vendor.

In the system, we implement the pattern matching mechanism and provide a user-friendly programming model for hardware vendors to easily specify patterns.
The programming model allows vendors to share the same infrastructure to match the sequence of operators that fit the specific instruction in their ISA; hence reducing engineering efforts in software development.
For instance, the following code snippet specifies a pattern of \texttt{Conv2D-Add-ReLU},

\begin{lstlisting}
def conv2d_pattern():
  data, weight, bias = wildcard(), wildcard(), wildcard()
  conv = is_op("nn.conv2d")(data, weight)
  bias_optional = conv.optional( \
    lambda x: is_op("nn.bias_add")(x, bias))
  return is_op("nn.relu")(bias_optional)
pattern_table = [(conv2d_pattern, "conv2d_pattern")]
\end{lstlisting}

\noindent where three \texttt{wildcard}s indicate the inputs of this pattern can be in any type, including input tensors of a model and the tensors computed by the parent operator.
\texttt{is\_op} is used to match the operator type such as \texttt{Conv2D}.
\texttt{optional} specifies that the \texttt{Add} is an optional node in this pattern, meaning that sequences with and without the \texttt{Add} are both matched.
Finally, line 7 assigns a name ``conv2d\_pattern'' to the pattern.

\begin{figure*}[tbh]
	\centering
    \includegraphics[width=0.8\linewidth]{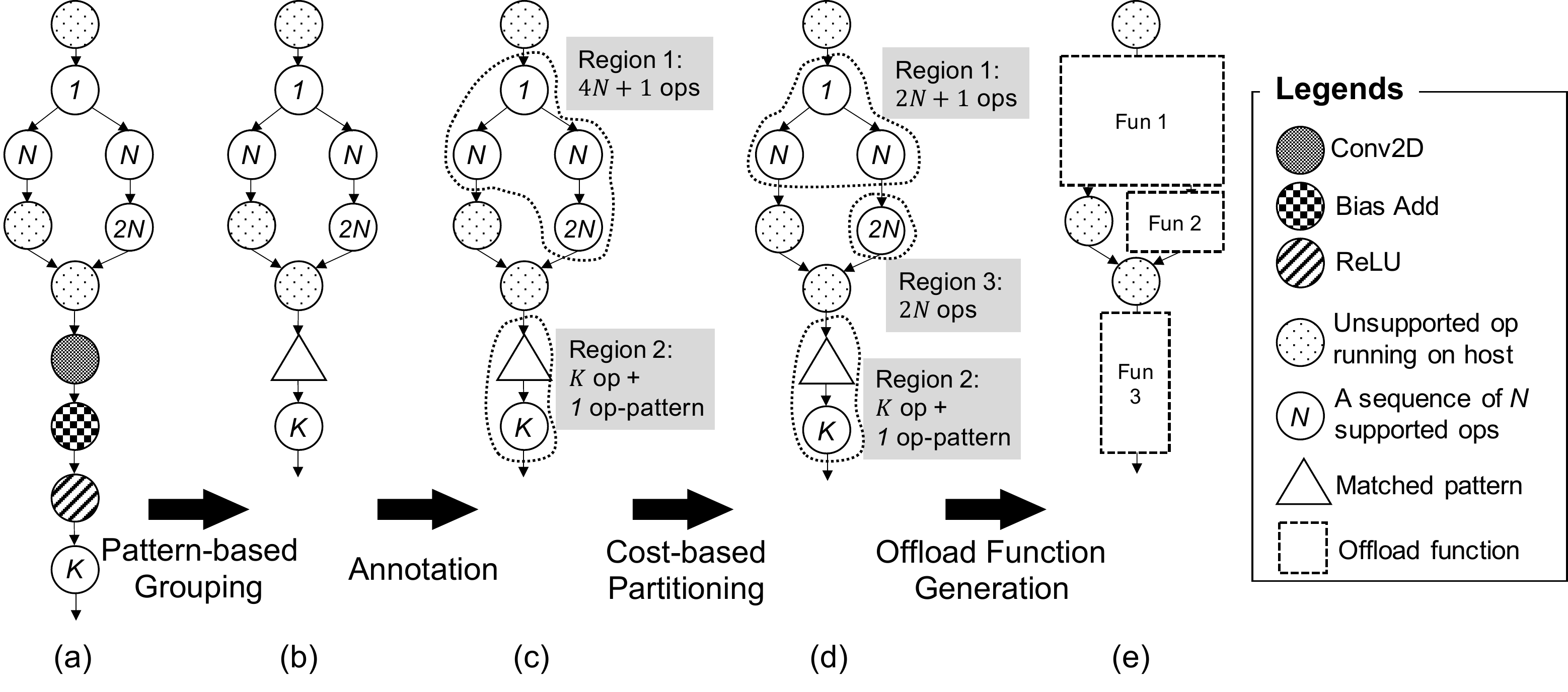}
    \caption{An illustrative example of graph partitioning.}
	\label{fig:partition}
\end{figure*}

By taking the above \texttt{pattern\_table}, we transform the model graph from \autoref{fig:partition}(a) to \autoref{fig:partition}(b).
As can be seen, the sequence has been replaced with a single graph node, which is a function including all three nodes.
The pattern name can be referred to in the code generation stage to map the operator sequence to the corresponding instruction supported by the accelerator.

\subsubsection{Annotation}
\label{sec:design_partition_op_annotate}
After grouping nodes based on the patterns, the next step is to use a programming model to easily specify a list of supported operators.
For example, the following code snippet registers a function to indicate that all \texttt{Conv2D} nodes with floating point data type should be annotated and offloaded to \texttt{myAccel}.
Since all attributes and input arguments of the node are accessible, one is able to specify any rules to determine whether a node can be offloaded or not.

\begin{lstlisting}
@register_op_attr("nn.conv2d", "codegen.myAccel")
def annotate_conv2d(expr):
    attrs, args = expr.attrs, expr.args
    if any([x.dtype != "float32" for x in args]):
        return False
    return True
\end{lstlisting}

By taking a set of annotation functions, we generate a number of regions, which can be potentially offloaded to the target accelerator, on the graph, as shown in \autoref{fig:partition}(c).
Note that in order to minimize the data transferring and kernel launching overheads, we greedily merge consecutive supported operators to one region and pass them together to the code generator.
In this particular example, we create two regions.
\textit{Region~1} contains $4N+1$ operators; while \textit{Region~2} contains $K$ operators as well as an operator pattern generated in \autoref{sec:design_partition_pattern}.

While we only focus on one target (i.e., \texttt{myAccel}) in this example, the proposed mechanism is applicable to multiple targets.
For example, if an operator can be offloaded to more than one targets, we allow users to specify the priority and then offload the operator to the one with the highest priority.
How to determine the target of each operator to achieve the optimal performance is an open research problem that is beyond the scope of this paper.

\subsubsection{Cost-based Partitioning}
\label{sec:design_partition_cost_partition}

Although greedily merging supported operators to maximize the region size is ideal, it might not be practical for some accelerators due to the resource constraints (e.g., on-chip memory size, number of compute units, etc.).
Therefore, in addition to the annotation, our framework offers a cost-based partitioning mechanism to split regions based on user-defined criteria.
In the illustrative example, we set the maximum number of operators to $3N$ as the criteria to split regions, and the result in the \textit{Region~1} in \autoref{fig:partition}(c) can be split to \textit{Region~1} and \textit{Region~3} in \autoref{fig:partition}(d).

In addition to the hardware resource constraints, offloading overhead is another concern.
Since offloading a region from host to the accelerator usually introduces data transfer and kernel invocation overheads, we should keep the region on the host if it has no time-consuming computation, such as MACs.
Accordingly, our framework accepts another layer of user specified criteria to fallback regions to the host (see \autoref{sec:study:trt} for an example of this).

Finally, each offload-able region is encapsulated into a separate function and labeled with a target attribute that indicates the backend on which it will be executed, as shown in \autoref{fig:partition}(e).
Accelerator-specific processing will be applied on these functions for further optimization before code generation, as we will discuss in the next subsection.




\subsection{Accelerator-Specific Processing}
\label{sec:design_backend_opt}
\begin{figure}[tbh]
	\centering
    \includegraphics[width=1.0\linewidth]{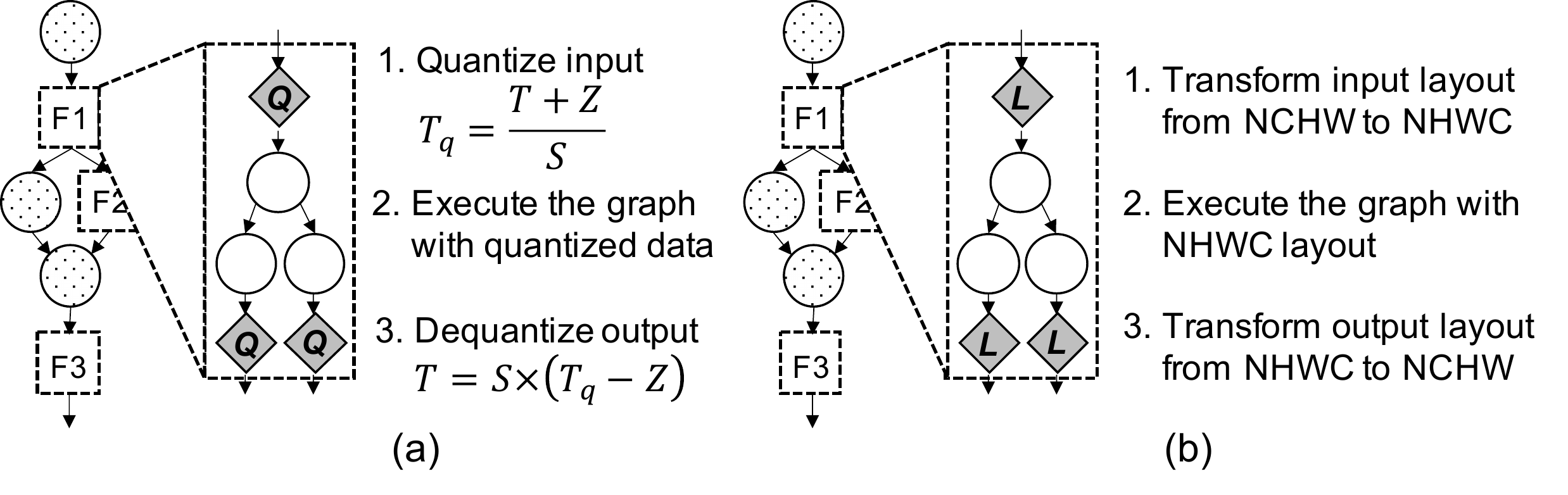}
    \caption{Examples of accelerator-specific processing. The 
diamond nodes in grey are inserted by the accelerator backend for special handling: (a) Quantization. \textit{\texttt{Q}} is a sequence of operators for (de)quantizating data. (b) Layout transformation. \textit{\texttt{L}} is an operator for transforming data layout.}
	\label{fig:proc}
\end{figure}

After the partitioning, a graph has been split to multiple regions that are handled by different backends.
The regions remain on the host can effectively leverage the standard optimizations from the existing deep learning compilers.
However, the regions offloaded to accelerators may require some hardware-dependent optimizations (e.g., fusion, substitution, layout transformation, quantization, etc.) that are not directly accessible from the deep learning compilers, as these optimizations are either proprietary or require specific hardware information.
We illustrate two common accelerator-specific processing with \autoref{fig:proc}.

\textbf{Quantization} is an important technique to reduce the resource utilization and energy consumption of DNN execution via quantizing the floating-point weights and activations into low-precision fixed-point numbers~\cite{han2016deep}.
Some accelerators can even only perform computations for fixed-point data.
Although hardware vendors could support pre-quantized models, which are quantized by deep learning frameworks with a calibration dataset, in their compiler tool chains to ensure that the aforementioned flow for accelerators is still applicable, it might not be practical to ask end-users to quantize every model by themselves before compilation and deployment.
With this framework, hardware vendors are able to enable ``partial quantization'', as shown in \autoref{fig:proc}(a).
Specifically, quantization and dequantization nodes (denoted as \textit{\texttt{Q}}) are inserted in the partitioned function to quantize between float-point tensors ($T$) and fixed-points tensors ($T_{q}$).
The constants $S$ and $Z$ are the scale and zero-point offset for the tensors, respectively, and they can be determined either at the compile time with the given calibration data, or at the run time with the real data.

\textbf{Layout transformation} is another important operation to make sure the input/output tensor layouts are computational friendly.
Since a hardware accelerator is usually optimized under a certain layout (e.g., \texttt{NCHW}\footnote{\texttt{NCHW} means \texttt{Conv2D} input data is organized in the order of (N)umber-(Channel)-(H)eight-(W)eight.} or \texttt{NHWC} data layout in \texttt{Conv2D}), it may have poor performance or even fail to perform computations on other layouts.
In this case, as shown in \autoref{fig:proc}(b), vendors could insert layout transformation nodes (denoted as \textit{\texttt{L}}) at the boundary of the partitioned functions to guarantee the input layouts are expected.

In summary, by maintaining accelerator-specific processes inside the partitioned function, our framework is capable of bringing the accelerators with certain requirements to the modern deep learning compilers.

\subsection{Code Generation}
\label{sec:design_codegen}
\begin{figure}[tbh]
	\centering
    \includegraphics[width=0.98\linewidth]{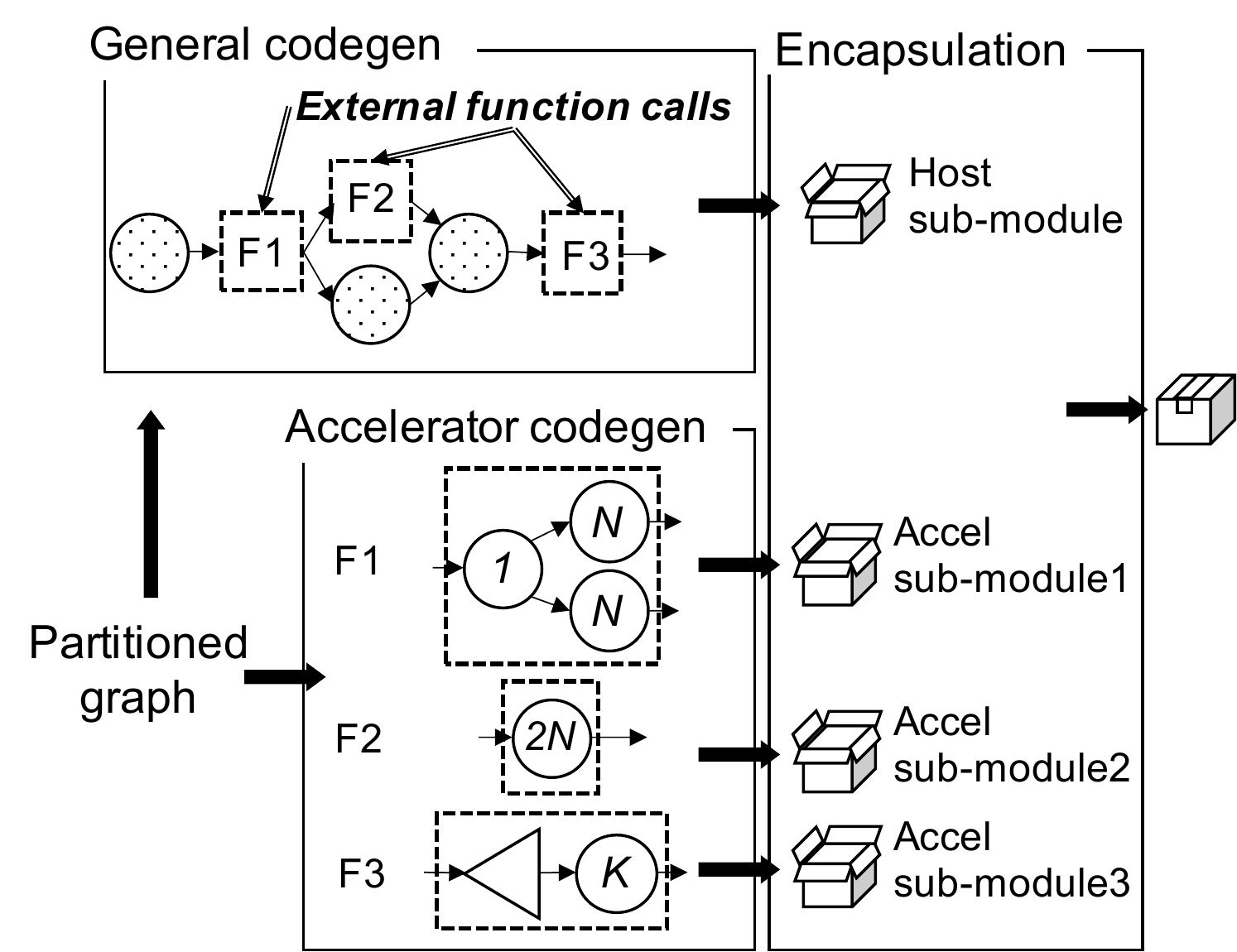}
    \caption{An illustrative example of the \system code generation flow. The input is the partitioned graph from \autoref{fig:partition}(e).}
	\label{fig:codegen}
\end{figure}

After partitioning, the final step of the compilation flow is the code generation.
\autoref{fig:codegen} depicts an overview of our framework's code generation flow.
Given a deep learning model graph with partitioned functions, the framework aims to generate a monolithic executable module for model inference.
The module includes 1) the model graph structure, and 2) the implementation of each graph node.
To achieve this goal, the framework generates a ``host sub-module'' by traversing the graph and invoking the corresponding code generation for each graph node.
When traversing to a node that remains on the host, we leverage the code generation in existing deep learning compilers, such as TVM~\cite{tvm} and XLA~\cite{xla}.
Most of them are capable of generating code for general devices (e.g., CPU and GPU).
When traversing to a node (i.e., a partitioned function) that is annotated with the specific target, we simply generate an external function call as a hook for kernel invocation at runtime.
Meanwhile, we invoke the accelerator specific code generation, which incorporates the hardware vendor specific code generation tools and compilation flow, to generate an ``accelerator sub-module'' for the partitioned function in that node.
The reason for separating the accelerator kernel implementations to different modules is to preserve the high flexibility for hardware vendors to integrate their compilation flows, as we will illustrate in the rest of this subsection.
Finally, one host sub-module and multiple accelerator sub-modules are then encapsulated together into a monolithic module as a heterogeneous blob.

The generated code of the partitioned function in accelerator sub-modules has to be represented in a certain format, so it can be consumed by the accelerators' execution engine at run time.
Our framework offers the following possible options to represent the generated code format.

\noindent\textbf{Option~1: Standard graph representation.} 
We choose JSON as the default graph format because it has gathered notable popularity in the deep learning community.
The code generation for this option simply translates a partitioned function into a JSON node with the equivalent information, including operator name, attributes, and data flow, etc.
This format is human readable and can be easily interpreted by runtime execution engines.
For instance, NVIDIA TensorRT~\cite{tensorrt} and Arm Compute Library~\cite{acl} leverage our JSON code generator to bridge the gap between our framework and their runtime engines.

\noindent\textbf{Option~2: Standard C code.} Although Option~1 is easy to implement and deploy, it requires a graph engine to include the implementations of all supported operators, which may result in the large binary size.
It is therefore not an ideal solution for some resource constrained accelerators that prefer to directly run the executable binary compiled from standard C code.

Hence, our framework offers a standard C code generator that can support accelerator's proprietary kernel libraries via emitting the kernel library function calls and linking them together with the host sub-module.
This solution eases code packaging because the host code is usually C compatible, which allows the library calls to be part of the host sub-module.
It implies that vendors do not have to customize their own runtime but can fully leverage the existing runtime system.

\noindent\textbf{Option~3: Custom graph representation.} Option~1 and Option~2 trade off simplicity, readability, and flexibility.
However, certain accelerators can only load the customized graph representation format which is different from the aforementioned ones.
For instance, ARM Ethos-N~\cite{armethosn} and Xilinx Vitis AI~\cite{xilinxvitis} have their own specialized stream formats to represent a neural network.
To accommodate such requirements, we also allow the customization of serialized code format.
In this scenario, hardware vendors could implement a set of unified APIs defined by our framework to specify the behavior of 1) compiling and serializing the generated code to a bit-stream so that it can be materialized with other sub-modules; 2) deserializing the bit-stream from the sub-module at runtime.

Until this point, the compiled and packed module for the deep learning model is ready.
In the next subsection, we present a lightweight runtime system to load the module and perform inferences.



\begin{figure*}[tbh]
	\centering
    \includegraphics[width=0.85\linewidth]{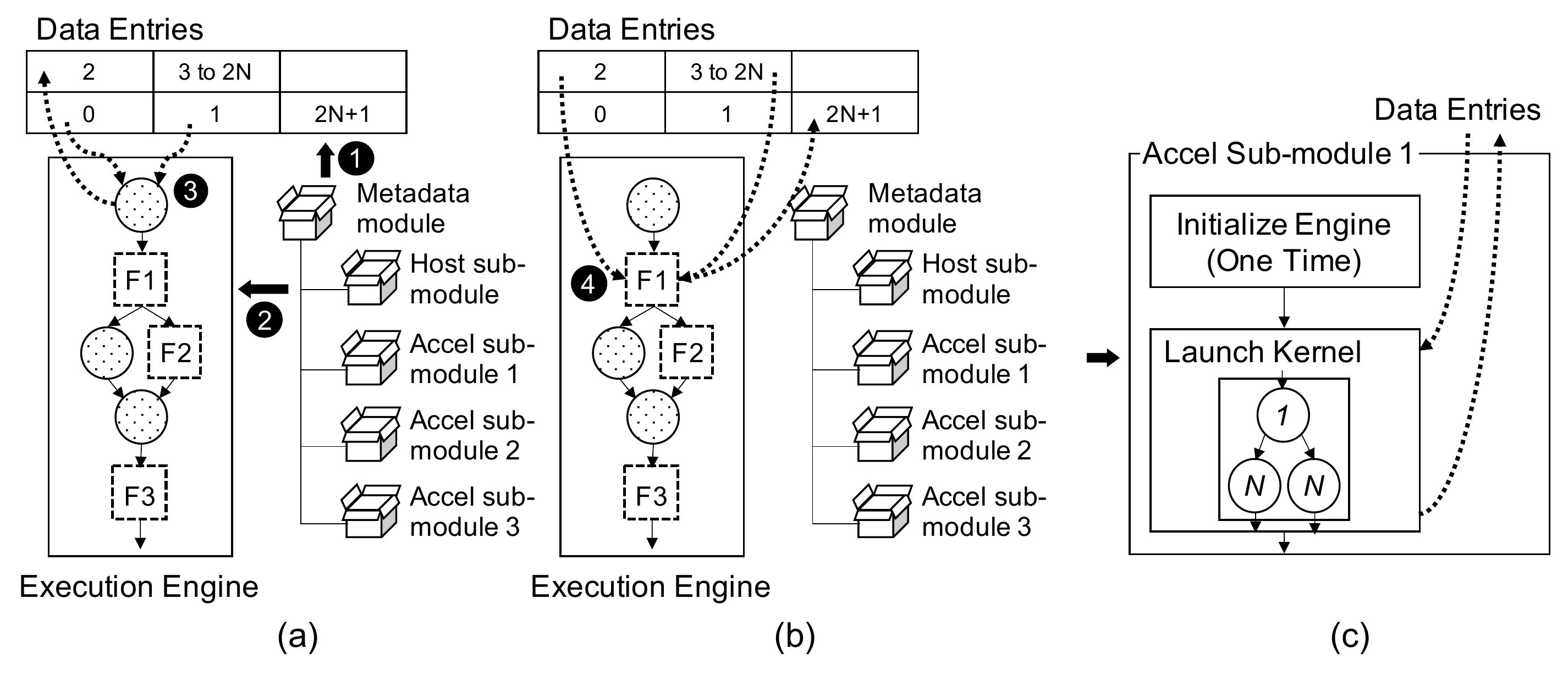}
    \caption{An illustrative example of runtime execution. (a) The process of loading a runtime module of the compiled model and executing the first operator on the host. (b) The process of executing an external function call on the accelerator. (c) The detail execution flow in the accelerator module for \texttt{F1}.}
	\label{fig:runtime}
\vspace{-0.2in}
\end{figure*}

\subsection{Runtime}
\label{sec:design_runtime}

Most deep learning compilers employ runtime systems to perform model inference.
As mentioned in the previous subsection, the compiler generates one or more runtime modules for the given deep learning model.
The runtime module is in charge of executing the model graph and dispatching the operators and subgraphs to the target platform.
The graph execution engine can be a simple dataflow graph visitor to deal with most convolutional neural networks (CNNs), or a virtual machine~\cite{shen2020nimble} to interpret the bytecode and handle dynamism and control flow presented in the modern models.
We design a unified runtime component to manage the host and accelerator-specific kernels in a hierarchical manner.
This component can be integrated into any runtime systems employed by existing deep learning compiler solutions.
We use an example illustrated in \autoref{fig:runtime} to introduce the runtime execution flow of our framework in detail.


\noindent\textbf{Initiate the metadata module.}
DNN models feature a number of weight parameters which are essentially constants during inference and should be included in the runtime modules with the model graph and generated kernels.
As illustrated in \autoref{fig:codegen}, for each deep learning model, our framework generates a single heterogeneous blob containing multiple runtime sub-modules for subgraphs on different target platforms.
It implies that the constants (i.e., weights) may be required by different modules, which leads to cumbersome management.
We propose a unified module, namely \textit{metadata module}, to handle the constants required by different modules.
As shown in \autoref{fig:runtime}(a), the metadata module is designed as a hierarchical module that contains all constants, the host sub-module \footnote{While CPU is normally used as the host, CUDA can also serve this role for TensorRT~\cite{tensorrt}.}, and the accelerator sub-modules.
Upon the initialization, metadata module loads the constants to the runtime data entries, which are a set of pre-allocated memory buffers on the host or the accelerator (\circled{1}).

In addition to the constants, data entries also maintain the model inputs and outputs as well as the intermediate results.
Since the partitioned functions are already external function calls, their intermediate results will not be maintained in the data entries, meaning that merging offloaded subgraphs can not only reduce the number of generated accelerator sub-modules, but also provide flexibility for hardware vendors to optimize the memory footprint.
For example, the vendor tools with a runtime engine such as NVIDIA TensorRT~\cite{tensorrt} are capable of reducing the memory consumption by fusing certain operators when building the runtime engine.


\noindent\textbf{Execute graph nodes on the host.}
When users invoke the inference process given the model input (e.g., an image), the host sub-module loads the model graph and initiates the execution engine on the host (\circled{2}).
The initiated execution engine then starts executing the graph nodes sequentially.
It is worth mentioning that since all inputs and outputs of each node are stored in the data entries with their IDs assigned during compilation, the runtime module can directly access the corresponding data entry to read the inputs or write the outputs.
Taking the first node in \autoref{fig:runtime}(a) as an example (\circled{3}), the host module reads the inputs from data entry 0 and 1, invokes the kernel to perform the computation, and writes the output to data entry 2.
The data entry 2 then becomes the input of the next node.

\noindent\textbf{Execute the graph nodes on the accelerator.}
Continuing the illustrative example in \autoref{fig:runtime}(b), the host execution engine now attempts to execute the node \texttt{F1}, which is an external function call to the accelerator (\circled{4}).
The execution details are depicted in \autoref{fig:runtime}(c).
As can be seen, the execution is composed of two steps.
The first step initiates the \texttt{F1} specific execution engine customized by hardware vendors on the completion of loading the module.
Depending on the implementation of the accelerator execution engine, the cost of this step may vary from micro seconds to even minutes (e.g., the execution engine performs just-in-time compilation).
THe system caches the execution engine after its initialization to eliminate this overhead for the later inferences. 
Given the initiated execution engine, the accelerator sub-module then reads the input from the assigned data entries, executes \texttt{F1} on the accelerator, and writes the output to another data entry.
%


\vspace{-0.2in}
\section{Case Study}
\label{sec:study}

We implemented our framework on top of the Apache TVM (version 0.7)~\cite{tvm}, an open source deep learning compiler infrastructure for its rich frontend parsers and hardware-agnostic optimization passes.
These benefits relieve the accelerator vendors from handling the framework specific details, which enables our work to focus on the design and implementation of the unified interfaces for accelerator integration.
Our technique is also applicable to other compilers.

We have worked with several hardware vendors to integrate their toolchains.
For example, the integration of Arm Ethos-N processor~\cite{armethosn} (2,405 LOC) and Xilinx Vitis-AI~\cite{xilinxvitis} with Deep Learning Processor Unit~\cite{dpu} (1,924 LOC) are enabled by our framework.
In addition, many vendor-specific software design kits (SDKs) such as NVIDIA TensorRT~\cite{tensorrt} (4,403 LOC), Arm Compute Library~\cite{acl} (2,188 LOC), Texas Instruments Deep Learning~\cite{tidl} (3,085 LOC), and CoreML~\cite{coreml} (840 LOC) are also available.

In this section, we dive into several representative cases and evaluate them in detail.
For each case, we first present their execution flow, starting from graph partitioning to the custom runtime.
Then we use a set of typical deep learning models from image classification and objection detection applications to evaluate the speedup achieved by the accelerators as well as the system overheads.
The chosen image classification models are ResNet-18~\cite{hecvpr2016resnet}, ResNet50\_v1b~\cite{hecvpr2016resnet}, Inception V3~\cite{inceptionv3}, DarkNet-53~\cite{darknet}, MobileNet V2~\cite{mobilenet}, and VGG19~\cite{liu2016ssd}, and the selected object detection models are SSD~\cite{ssd} with MobileNet and ResNet-34 as the backbone network and Faster R-CNN~\cite{fastrcnn} due to their popularity in edge applications.
Since some accelerators such as Ethos-N are not publicly available yet for evaluation, we study NVIDIA JetSon Xavier and Xilinx Vitis AI in detail in this section.
For each case, we seek to explore the following questions.

\begin{itemize}
\itemsep0cm
    \item What is the obtainable speedup of the models compared to the selected baseline on the studied platforms?
    \item How many operators can be offloaded to the target platform?
    \item What is the percentage of the MACs the offloaded graphs account for?
    \item How much is the offloading overhead?
\end{itemize}

\subsection{NVIDIA Jetson AGX Xavier GPUs}
\label{sec:study:trt}
\begin{table*}[tbh]
\centering
\caption{End-to-end performance on NVIDIA Jetson Xavier with TensorRT with full/half precision kernels. The baselines are full-precision models compiled by the TVM builtin CUDA code generation.}
\label{tbl:trt-all}
\resizebox{0.7\textwidth}{!}{%
\begin{tabular}{ccccccc}
\hline
\multirow{2}{*}{Model} & \multicolumn{2}{c}{Offload Ratio (\%)} & \multicolumn{2}{c}{Full-Precision} & \multicolumn{2}{c}{Half-Precision} \\ \cline{2-7} 
                          & Node  & MAC   & Latency (ms) & Speedup & Latency (ms) & Speedup \\ \hline
ResNet-18                 & 100\% & 100\% & 4.12         & 1.07    & 2.14         & 2.06    \\ \hline
ResNet-50 v1b/v2          & 100\% & 100\% & 9.66         & 1.53    & 4.12         & 3.58    \\ \hline
Inception V3              & 100\% & 100\% & 16.13        & 2.00    & 6.10         & 5.30    \\ \hline
DarkNet-53                & 100\% & 100\% & 12.81        & 3.73    & 5.43         & 8.81    \\ \hline
MobileNet V2 1.0          & 100\% & 100\% & 3.32         & 1.19    & 2.22         & 1.78    \\ \hline
VGG19                     & 100\% & 100\% & 24.26        & 0.92    & 9.78         & 2.29    \\ \hline
SSD 512 MobileNet 1.0     & 36\%  & 100\% & 18.43        & 2.24    & 11.06        & 3.73    \\ \hline
SSD 300 ResNet-34         & 41\%  & 100\% & 34.31        & 48.44   & 15.29        & 108.68  \\ \hline
Faster R-CNN ResNet50 v1b & 63\%  & 100\% & 404.51       & 6.29    & 99.72        & 25.53   \\ \hline
\end{tabular}%
}
\end{table*}
We start the case study with the NVIDIA Jestson AGX Xavier GPU, which is the latest edition to the Jetson platform as of the writing of this paper.
Jetson AGX Xavier contains an integrated 512-core Volta GPU with Tensor Cores, two Deep Learning Accelerators (DLAs), and a 8-core NVIDIA Carmel ARMv8.2 CPU.
It is mainly designed for robots, drones, and other autonomous machines~\cite{nvidiaxavier}.

In this study, we have integrated TensorRT in our framework using the proposed flow.
The entire integration was implemented in C++ and Python with $\sim$4.4K LOC.
Since the TensorRT supports both 32-bit (full-precision) and 16-bit (half-precision) floating point kernels, our integration is able to compile either of them based on users.
When running half-precision TensorRT kernels with full-precision models, we leverage the partial quantization mechanism introduced in \autoref{sec:design_backend_opt} to cast full-precision input data to half-precision on the fly.

To investigate the performance gain of partitioning the graph and leveraging TesnorRT, we apply the TVM builtin CUDA code generation for an entire full-precision model as the baseline.
Note that although TVM has an auto-tuning framework named AutoTVM~\cite{chen2018learning} to tune a given model on the target device, we do not use AutoTVM to tune the baselines.
The reason is that AutoTVM leverages on-device measurements to guide the search, and every operator needs at least 4,000 trials to achieve decent performance, which usually takes more than 3 hours on the device with limited CPU computation resources such as Jetson Xavier.
As a result, tuning an SSD model with ResNet-34 backbone on Jetson Xavier may take a week.

\autoref{tbl:trt-all} shows the end-to-end latency and speedup on NVIDIA Jetson Xavier using our framework with TensorRT over the baseline.
As shown in the table, we are able to offload all operators of image classification models to TensorRT.
For object detection models such as SSD and Faster R-CNN, we can only offload a fraction of their operators, because the control flow operators (e.g., \texttt{If}, \texttt{For}) and some other operators (e.g., \texttt{arange}) are not supported.
However, the offload ratio of MACs\footnote{Usually only \texttt{Conv2D} and \texttt{Dense} have MAC computations in DNNs.} are still 100\% for both models, meaning that we have already offloaded all time-consuming computations to TensorRT.
In addition, we can see from \autoref{tbl:trt-all} that the speedups that TensorRT integration achieved are notable on some models but moderate on a few other models such as VGG19.
This is because TensorRT leverages hand-crafted kernels which is not scalable to all possible workloads.

On the other hand, the speedup achieved by offloading half-precision kernels is more significant across virtually all studied benchmarks.
Compared to the latencies with full-precision TensorRT kernels, the latencies with half-precision TensorRT kernels achieve on average 2.36$\times$ speedup.
Intuitively, half-precision computation is expected to achieve up to 2$\times$ speedup over full-precision due to less computation, but since TensorRT enables the use of Tensor Cores only with half-precision computation, the speedup of half-precision could be more than 2$\times$ with TensorRT.

We note that the above latency numbers do not include the one-time TensorRT engine initialization overhead.
Since we choose to generate the subgraphs in the JSON format (Option~1 in \autoref{sec:design_codegen}) for TensorRT code generation, a TensorRT engine has to be built for each subgraph when we initialize the runtime module.
According to our experiments, the initialization time of the execution engine for each kernel on NVIDIA Jetson Xavier could range from 8 to 62 seconds.
Since the runtime module will cache the built TensorRT engine, the inference latency will not be affected.
With the built engine, the kernel invocation overhead is negligible during the inference. 
Moreover, there is no data transferring cost either during the inference due to the use of CUDA as the host.
It implies that both the host (CUDA) and the accelerator (TensorRT) access the same GPU memory; hence eliminating the need of copying tensors from the host to the accelerator memory.
In summary, the TensorRT integration trades the inference overhead with the long initialization time.

\begin{table}[tbh]
\centering
\caption{Subgraph number and latency comparison with and without cost-based partitioning of TensorRT integration on NVIDIA Jetson Xavier.}
\label{tbl:trt-cost}
\resizebox{0.48\textwidth}{!}{%
\begin{tabular}{ccccc}
\hline
\multirow{2}{*}{Model}    & \multicolumn{2}{c}{Without} & \multicolumn{2}{c}{With} \\ \cline{2-5} 
 &
  \begin{tabular}[c]{@{}c@{}}Total\\ Subgraph \#\end{tabular} &
  \begin{tabular}[c]{@{}c@{}}Latency\\ (ms)\end{tabular} &
  \begin{tabular}[c]{@{}c@{}}Total\\ Subgraph \#\end{tabular} &
  \begin{tabular}[c]{@{}c@{}}Latency\\ (ms)\end{tabular} \\ \hline
ResNet-18                 & 1          & 4.12           & 1        & 4.12          \\ \hline
ResNet-50 v1b/v2          & 1          & 9.66           & 1        & 9.66          \\ \hline
Inception V3              & 1          & 16.13          & 1        & 16.13         \\ \hline
DarkNet-53                & 1          & 12.81          & 1        & 12.81         \\ \hline
MobileNet V2 1.0          & 1          & 3.32           & 1        & 3.32          \\ \hline
VGG19                     & 1          & 24.26          & 1        & 24.26         \\ \hline
SSD 512 MobileNet 1.0     & 6          & 48.13          & 1        & 18.43         \\ \hline
SSD 300 ResNet-34         & 6          & 79.30          & 1        & 34.31         \\ \hline
Faster R-CNN ResNet50 v1b & 21         & 407.15         & 2        & 404.51        \\ \hline
\end{tabular}%
}
\end{table}

Finally, we study the impact of subgraph numbers in a model.
\autoref{tbl:trt-cost} shows the number of subgraphs before and after the cost-based partitioning and the end-to-end inference latency.
In the TensorRT integration, we simply set the cost function to $CalcMAC(graph) > 0$ to fallback all subgraphs without any MACs to the host to avoid overheads.
For classic CNN models (e.g., ResNet, MobileNet), since we are able to offload the entire model to TensorRT, cost-based partitioning is not necessary.
On the other hand, cost-based partitioning works quite well for the object detection models, such as SSD and Faster R-CNN.
These models contains many offload-able data processing operators (e.g., \texttt{transpose}, \texttt{maximum}, \texttt{reshape}) that cannot be grouped with other compute intensive operators such as \texttt{Conv2D} due to the unsupported control flow in between.
As a result, many subgraphs in these models have no compute intensive operators and cannot achieve decent speedup by TensorRT, so keeping them on the host device could reduce the overhead.

We further dive into the overheads of six subgraphs before cost-based partitioning in SSD-512 with MobileNet backbone.
Except for the subgraph that includes an entire backbone network, other subgraphs do not have any MAC computations so their kernels may not have speedup over the baseline (ranging from 0.48$\times$ to 1.15$\times$).
Even worse, we still need to pay for the initialization time of their execution engines, which ranges from 0.4 to 19 seconds.
This illustrates that cost-based partitioning is practical to control the overheads and achieve overall better performance.



\subsection{Xilinx Edge and Cloud FPGAs}
The second study is Xilinx Vitis AI~\cite{xilinxvitis}, Xilinx’s development stack for deep learning model inference on both Xilinx edge devices and Alveo accelerator cards.
Xilinx Zynq UltraScale+ FPGA and U250 FPGA are chosen to evaluate the integration of Vitis AI through our framework.
The former is built on TSMC 16nm FinFET+ process technology and compatible with the Zynq-7000 SoC.
It features a 64-bit 8-core ARM Cortex-A53 processor, a dual-core ARM Cortex-R5 real-time processor, and an ARM Mali$^{TM}$-400MP Graphics Processor.
The latter is a data center accelerator card that built on the Xilinx 16nm UltraScale$^{TM}$ architecture.
It is able to deliver up to 90$\times$ higher performance than CPUs on workloads, such as machine learning inference and video transcoding, at much lower cost~\cite{xilinxu250}.

Vitis AI is integrated into the system flow with $\sim$2K LOC in C++ and Python.
The integration has a custom bit-stream format to represent the partitioned subgraphs, and the custom runtime module with Vitis-AI is capable of interpreting the bit-stream format and dispatching operators to the deployed Xilinx DPUs.
In addition, since Xilinx DPUs only focus on fixed-point computations to fully utilize the power of FPGAs, this integration incorporates the customized partial quantization.
In particular, the customized Vitis-AI runtime feeds the data to the Xilinx's quantizer to calculate the $T$ and $Z$ in \autoref{fig:proc}.
Consequently, although the rest of evaluations are based on the models with full-precision floating points, the part that offloaded to the FPGAs still performs fixed point computation.

\begin{table}[tbh]
\centering
\caption{End-to-end performance of full precision models on Xilinx U250 cloud FPGA. The baseline is compiled by the TVM builtin LLVM code generation and runs on Intel(R) Xeon(R) Gold 6252 CPU.}
\label{tbl:xilinx-cloud}
\resizebox{0.48\textwidth}{!}{%
\begin{tabular}{ccccc}
\hline
\multirow{2}{*}{Model} &
  \multirow{2}{*}{\begin{tabular}[c]{@{}c@{}}Latency\\ (ms)\end{tabular}} &
  \multirow{2}{*}{Speedup} &
  \multicolumn{2}{c}{Offload Ratio (\%)} \\ \cline{4-5}
                          &         &       & Node  & MAC  \\ \hline
ResNet-18                 & 2.66    & 3.91$\times$ & 96\% & 99\%   \\ \hline
ResNet-50 v1b/v2          & 5.59    & 4.48$\times$ & 98\% & 99\%   \\ \hline
Inception V3              & 7.51    & 4.14$\times$ & 99\% & 99\%   \\ \hline
DarkNet-53                & 6.78    & 5.07$\times$ & 98\% & 100\%  \\ \hline
VGG19                     & 15.82   & 5.52$\times$ & 80\%  & 99\%  \\ \hline
SSD 300 ResNet-34.        & 43.21   & 2.42$\times$ & 33\%  & 100\% \\ \hline
Faster R-CNN ResNet50 v1b & 1,080 & 1.08$\times$ & 46\% & 20\%     \\ \hline
\end{tabular}%
}
\end{table}

\autoref{tbl:xilinx-cloud} details the speedup of the studied models on the Xilinx U250 cloud FPGA\footnote{MobileNet and SSD 512 MobileNet are skipped in this evaluation because depthwise Conv2D is not widely used in the cloud and not supported by U250 DPU.}.
Since the accelerated computing instances on public clouds, such as Amazon EC2 F1 instances~\cite{amazon-f1}, are usually based on general computing instances with a high-end CPU, the baseline latency of this study is measured on the Intel Xeon Gold 6252 CPU, which has 96 logical cores (2 sockets$\times$24 cores$\times$2 threads/core), with TVM LLVM code generation.
As can be seen in \autoref{tbl:xilinx-cloud}, the offload ratio of operators does not achieve 100\% due to unsupported operators (e.g., \texttt{strided\_slice}, \texttt{split}, \texttt{non maximum suppression}).
However, there are still significant performance gains by fully offloading the MAC computations to FPGAs for almost all models except for Faster R-CNN.
This is because the current Vitis-AI integration only supports one kernel in a model, so it only offloads the backbone ResNet-50 to the FPGA for Faster R-CNN, but the R-CNN takes a significant amount of time.

\begin{table}[tbh]
\centering
\caption{End-to-end performance of full precision models on Xilinx Zynq UltraScale+ edge FPGA. The baseline is compiled by the TVM builtin LLVM code generation and runs on ARM Cortex-A53 CPU.}
\label{tbl:xilinx-edge}
\resizebox{0.48\textwidth}{!}{%
\begin{tabular}{ccccc}
\hline
\multirow{2}{*}{Model} &
  \multirow{2}{*}{\begin{tabular}[c]{@{}c@{}}Latency\\ (ms)\end{tabular}} &
  \multirow{2}{*}{Speedup} &
  \multicolumn{2}{c}{Offload Ratio (\%)} \\ \cline{4-5}
                         &          &       & Node  & MAC   \\ \hline
ResNet-18                & 5.83     & 34.05$\times$ & 96\% & 100\% \\ \hline
ResNet-50 v1b/v2         & 14.7     & 44.51$\times$ & 98\% & 99\%  \\ \hline
Inception V3             & 19.47    & 77.34$\times$ & 98\% & 100\% \\ \hline
DarkNet-53               & 18.29    & 75.01$\times$ & 98\% & 100\% \\ \hline
MobileNet V2 1.0         & 4.72     & 19.11$\times$ & 99\% & 100\% \\ \hline
VGG19                    & 132.78   & 10.22$\times$ & 80\%  & 99\% \\ \hline
SSD 512 MobileNet 1.0    & 352.69   & 5.47$\times$  & 28\% & 100\% \\ \hline
SSD 300 ResNet-34        & 467.91   & 12.7$\times$  & 34\% & 100\% \\ \hline
Faster R-CNN ResNet50 & 49,016 & 1.29$\times$  & 46\% & 20\%.      \\ \hline
\end{tabular}%
}
\end{table}

In \autoref{tbl:xilinx-edge}, we report the results on Xilinx Zynq UltraScale+ edge FPGA.
As can be seen, the offload ratio is not exactly the same compared with \autoref{tbl:xilinx-cloud} due to the different hardware resource constraints.
However, the achieved speedup over the baseline on ARM Cortex-A53 CPU is more significant.
This illustrates that the accelerators on edge are more beneficial than cloud, since the edge CPU usually has limited resources.

\begin{figure}[tbh]
	\centering
    \includegraphics[width=\linewidth]{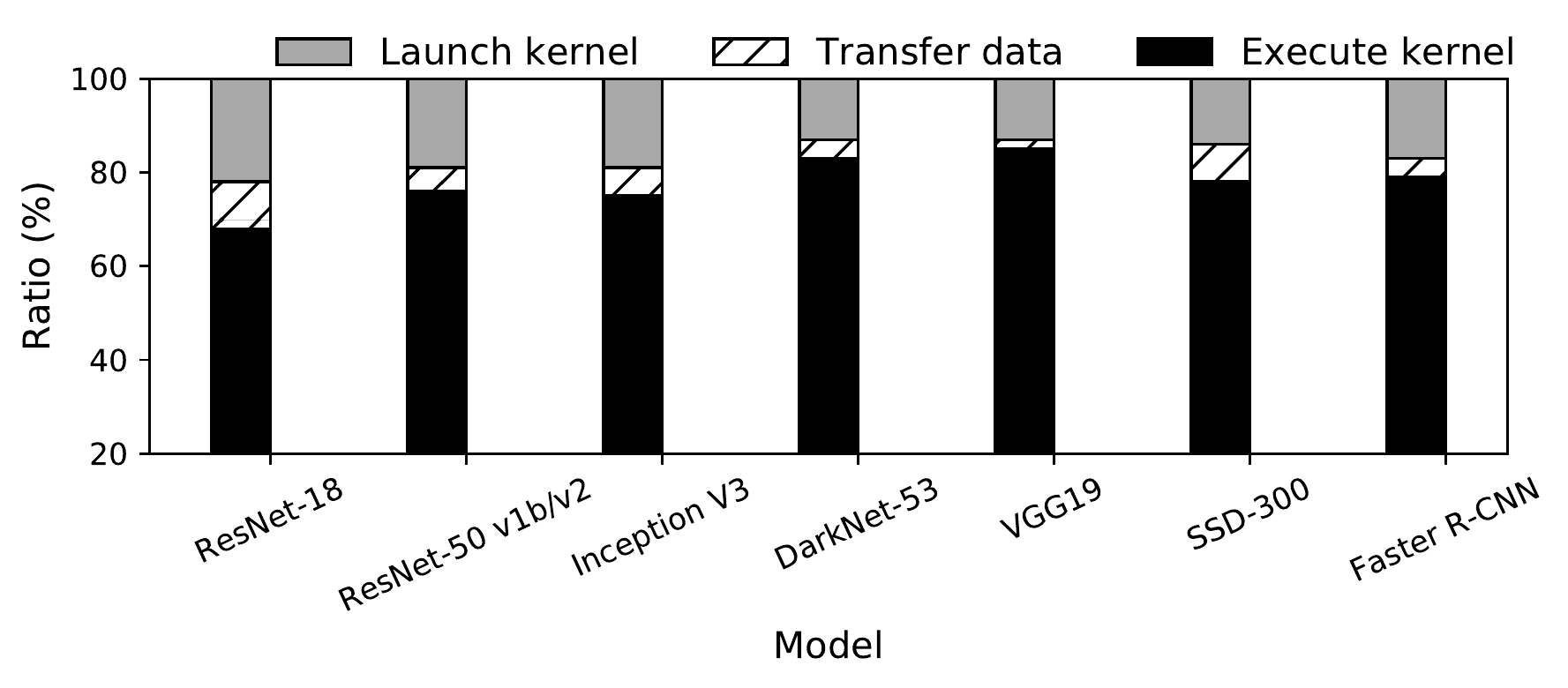}
    \caption{The breakdown of kernel execution on Xilinx U250 cloud FPGA. The data transfer and kernel invocation time take on average 6\% and 20\%, respectively.}
	\label{fig:xilinx-cloud}
\end{figure}
\begin{figure}[tbh]
	\centering
    \includegraphics[width=\linewidth]{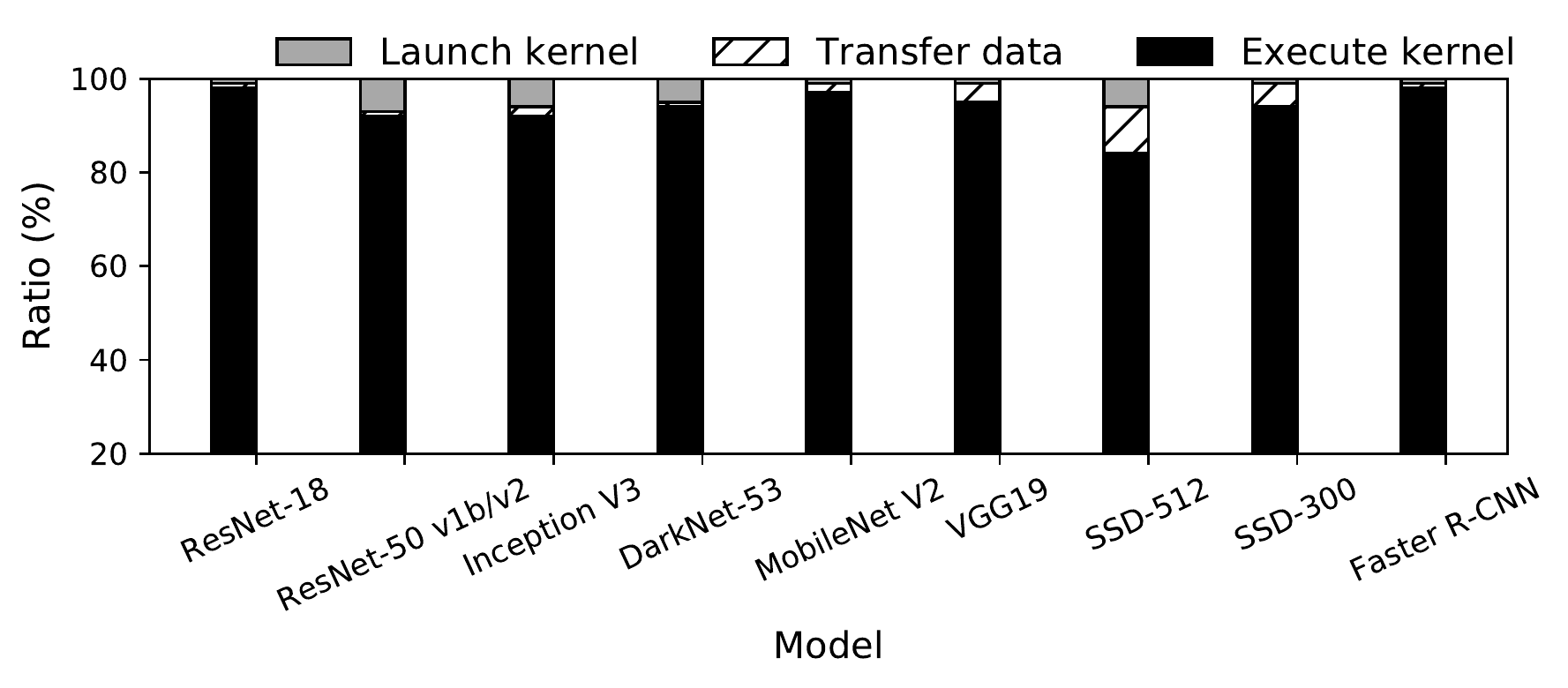}
    \caption{The breakdown of kernel execution on Xilinx Zynq UltraScale+ edge FPGA. Both data transfer and kernel invocation time take on average 3\%.}
	\label{fig:xilinx-edge}
\vspace{-0.2in}
\end{figure}

Finally, we evaluate the offloading overhead by studying the execution time breakdown of the investigated models on both Xilinx cloud and edge FPGAs in \autoref{fig:xilinx-cloud} and \autoref{fig:xilinx-edge}.
Unlike TensorRT that builds an execution engine for each subgraph, Xilinx Vitis-AI runtime is capable of executing the graph directly on its DPU.
As a result, it does not have the engine initialization overhead.
Other than that, in \autoref{fig:xilinx-cloud}, the kernel execution time dominates the overall latency in most cases.
The data transfer time takes only 6\% on average, and the kernel invocation time accounts for roughly 20\%.
The relatively high kernel invocation time includes data formatting and partial quantization as mentioned in \autoref{sec:design_backend_opt}.

Finally, in \autoref{fig:xilinx-edge}, both the data transfer and kernel invocation overheads take only on average 3\% except for the kernel in SSD-512 due to the large number of subgraph outputs (i.e., 18) that result in large size of data to be transferred.
\vspace{-2em}
\section{Related Work}
\label{sec:relw}

In this section, we survey a subset of representative efforts on deep learning compilers (DLC) for customized accelerators.

\noindent\textbf{General DLC with custom backend support: }
Other than our work, there are a few existing general DLCs \cite{onnxruntime,rotem2018glow} for hardware accelerator vendors to integrate.
All of them provide a mechanism to partition a deep learning model and partially offload the model to the accelerator.
In particular, ONNX runtime~\cite{onnxruntime} is a runtime system that interprets and executes a deep learning model in ONNX (Open Neural Network eXchange) format.
Vendors could provide a list of supported operators as well as customize an ONNX runtime module for their accelerators, and the runtime operator dispatcher will offload the supported operators to the custom runtime running on the accelerator.
However, interpreting a deep learning model at the runtime loses the opportunity of optimizing the model graph with the consideration of hardware accelerators.
Accordingly, ONNX runtime executes the graph operator-by-operator, which may introduce unnecessary data transfer overheads.
Another open source DLC, Glow~\cite{rotem2018glow}, provides a compiler backend in addition to the runtime, so that vendors could customize a graph partitioner and code generator that partitions the graph and generates the accelerator code for supported subgraphs.
Unfortunately, Glow only provides a CPU backend as the fallback solution, and it lacks of details in \cite{rotem2018glow} such as how each subgraph is compiled and serialized.

\noindent\textbf{Accelerator-specific DLC: }
Numerous specialized architectures for deep learning acceleration have been proposed in recently years.
To achieve the peak performance for the given deep learning models on these architectures, the developers usually present a few specialized intrinsic instructions or a completely new ISA.
In both cases, they require a code generator to generate the executable code with the instruction from the given deep learning model.
For example, the accelerator PUMA (Programmable Ultra-efficient Memristor-based Accelerator) \cite{ankit2019puma} comes up with a C++ runtime that parses the ONNX graph model and runs it with the proposed ISA.
The authors in \cite{fujiki2018memory} proposed an in-memory data parallel processor, and implement their compiler based on TensorFlow. 
The authors of Field Programmable Synapse Array (FPSA)~\cite{ji2019fpsa} implemented a compiler system that accepts the deep learning model graph and maps to the FPSA architecture.
All compilers of the above architectures could be potentially built upon our framework to not only reduce the efforts of implementation as well as maintenance, but make their accelerators catch up with the state-of-the-art deep learning models and frameworks.

On the other hand, another direction maintains a DLC for a deep learning acceleration friendly ISA.
For example, VTA~\cite{moreau2018vta} provides a full compiler stack and supports various deep learning frameworks.
NVIDIA also maintains a compiler for its deep learning accelerators~\cite{nvdla}.
As long as the accelerator developers follow the ISA and I/O ports, their accelerators are naturally supported by the official DLCs.
However, it restricts the flexibility of the developers when designing accelerators. 

Instead of proposing an architecture for application-specific integrated circuit (ASIC), some accelerator developers proposed a parameterized neural network architecture that fits field programmable gate arrays (FPGAs), so that the architecture can be further customized for each deep learning model~\cite{sohrabizadeh2020autodse, sharma2016dnnweaver,ma2018alamo,lai2019heterocl,cong2018polysa}.
Specifically, they proposed an architecture ``template'' with a set of configurable parameters such as process unit (PE) number and buffer sizes.
Based on the proposed network architecture, a compiler is implemented to generate an FPGA accelerator for the given neural network.
Since these accelerator-specific compilers also include common optimizations (e.g., constant folding, expression simplification, layout transformation, and quantization), the implementation efforts can be significantly reduced if they are built on our framework.

\vspace{-1.5em}
\section{Conclusion} \label{sec:concl}
This paper proposed a unified framework to allow accelerator vendors to easily integrate their codegen and runtime systems to state-of-the-art deep learning compilers.
This enables vendors to only focus on the development of the in-house code generator and greatly benefit from the development of standard compiler and deep learning optimization techniques in existing deep learning compilers.
We demonstrated a series of design and implementation mechanisms in graph partitioning, accelerator-specific processing, code generation, and the runtime systems to realize the framework.
Multiple commercial accelerator vendors have integrated their compiler stack to our framework.
NVIDIA Jetson Xavier and Xilinx Vitis-AI are studied in detail to illustrate the simplicity of the integration and their performance gains over the baselines.



\bibliographystyle{plain}
\interlinepenalty=10000
\bibliography{ref}

\end{document}